\documentclass{ifacconf}

\makeatletter
\let\old@ssect\@ssect 
\makeatother

\usepackage{graphicx}      
\usepackage{natbib}        
\usepackage{amsmath}
\usepackage{amsfonts}
\usepackage{booktabs}
\usepackage{color,xcolor}
\usepackage[overload]{textcase}
\usepackage{hyperref}

\newcommand{\new}[1]{%
    #1
}

\makeatletter
\def\@ssect#1#2#3#4#5#6{%
  \NR@gettitle{#6}
  \old@ssect{#1}{#2}{#3}{#4}{#5}{#6}
}
\makeatother

\begin{document}
\begin{frontmatter}

\title{Deep-learning-based Early Fixing for Gas-lifted Oil Production Optimization: Supervised and Weakly-supervised Approaches\thanksref{footnoteinfo}}

\thanks[footnoteinfo]{This research was partly funded by Petr\'oleo Brasileiro S.A. under grant 2018/00217-3, FAPESC under grant 2021TR2265, CNPq and CAPES.}

\author[1]{Bruno M. Pacheco} 
\author[1]{Laio O. Seman} 
\author[1]{Eduardo Camponogara}

\address[1]{Department of Automation and Systems Engineering, Federal University of Santa Catarina (UFSC), Florianópolis, Brazil (e-mail: bruno.m.pacheco@grad.ufsc.br, laio@ieee.org, eduardo.camponogara@ufsc.br).}

\begin{abstract}                
Maximizing oil production from gas-lifted oil wells entails solving Mixed-Integer Linear Programs (MILPs).
As the parameters of the wells, such as the basic-sediment-to-water ratio and the gas-oil ratio, are updated, the problems must be repeatedly solved.
Instead of relying on costly exact methods or the accuracy of general approximate methods, in this paper, we propose a tailor-made heuristic solution based on deep learning models trained to provide values to all integer variables given varying well parameters, early-fixing the integer variables and, thus, reducing the original problem to a linear program (LP).
We propose two approaches for developing the learning-based heuristic: a supervised learning approach, which requires the optimal integer values for several instances of the original problem in the training set, and a weakly-supervised learning approach, which requires only solutions for the early-fixed linear problems with random assignments for the integer variables.
Our results show a runtime reduction of 71.11\% 
Furthermore, the weakly-supervised learning model provided significant values for early fixing, despite never seeing the optimal values during training.
\end{abstract}


\end{frontmatter}

\section{Introduction}

The maximization of oil production in an offshore platform is a challenging problem due to the physical models' complexity and various operational constraints~\citep{luguesi_2023,muller_short-term_2022}.
Furthermore, the variations in the oil wells and the multitude of technologies that can be employed account for the many configurations this problem presents.
Therefore, it becomes necessary to rely on optimization models to reach optimal conditions for the operation.


Due to the nonlinearities of the oil output stream from gas-lifted wells, the formulation of the optimization model is not straightforward.
The approach of \cite{muller_short-term_2022} uses a piecewise linear model of the wells and formulates the problem as a Mixed-Integer Linear Program (MILP).
More specifically, the relationship between the liquid production of the wells and the lift-gas injection, together with the wellhead pressure, is modeled as a piecewise linear function.
The piecewise-linear functions are defined by studying the behavior of the relations through simulation runs or field tests and defining breakpoints between which the relation is considered to be linear.
Special Ordered Set of type 2 (SOS2) constraints are then used to define the operational region for each constraint, which defines the active boundaries for a region in the state space in which the function approximation is linear.
The final model is linear with binary variables and SOS2 variable sets.

Solving MILPs exactly requires algorithms such as branch-and-bound and branch-and-cut \citep{lee_integer_2009}.
These algorithms usually require long runtimes due to the many iterations of solving linear relaxations of the original problem.
One can use approximate methods instead of relying solely on exact algorithms to optimize oil production.
Such methods can improve the efficiency of the optimization process but may provide suboptimal solutions.
Derivative-free methods like genetic algorithms, simulated annealing, and particle swarm optimization can also be used to find near-optimal solutions efficiently~\citep{seman_derivative-free_2020}.
Another approximate method is to early fix the variables, reducing the dimension of the problem.
In the case of MILPs, one can develop a heuristic to fix all integer variables, reducing the problem to a linear program, which can be solved very efficiently with algorithms such as the simplex.
As pointed out by \cite{bengio_machine_2021}, such heuristics can be very hard to handcraft, which makes machine learning (ML) models natural candidates for the task.

In this paper, we propose two deep-learning approaches to reduce the runtime of the gas-lifted oil production MILP through early fixing.
One is a \textit{supervised learning approach} that requires that the optimal set of binary variables is known for each problem instance; instances of the problem must be solved exactly to build the training data.
With this data, the model is then trained to provide the optimal binary variables given the parameters of the MILP.
The other is a \textit{weakly-supervised approach} that requires just the solutions to the early-fixed problem, i.e., after fixing the integer variables to any (binary) value.
In other words, the training data is generated from solving linear problems, resulting from randomly fixing the integer variables in the original MILP.
\new{%
To the best of our knowledge, this is the first work to successfully implement learning-based heuristics to speed up the solution of oil production optimization.
}


\subsection{Related work}


In this section, we provide a brief overview of related work in approximate methods for MILP, focusing on deep-learning-based methods.

The use of heuristics in MILP solvers is common.
For instance, the SCIP solver \citep{vigerske_scip_2018} uses primal heuristics to find feasible solutions.
However, recent studies have highlighted the development of heuristics based on ML techniques.
Particularly, \cite{bengio_machine_2021} have suggested that the potential benefits of using ML include reduced computational time and improved solution quality.

\cite{ding_accelerating_2019} presented a learning-based heuristic to accelerate the solution-finding process.
The authors propose a Graph Neural Network (GNN) model that predicts solution values for branching.
The developed heuristic is used to guide a branch-and-bound tree search.

\cite{li_learning_2022} train a learning-based heuristic for early fixing MILPs within an Alternating Direction Method of Multipliers (ADMM) algorithm.
By formulating the early fixing as a Markov decision process, the authors use reinforcement learning to train the heuristic.
The authors showed that the proposed heuristic significantly improves the solving speed and can even improve the solution quality.

\cite{pacheco2023graph} explore using GNNs as early-fixing heuristics for the Offline Nanosatellite Task Scheduling (ONTS) problem.
In this direction, the authors implement a GNN-based solution that uses bipartite graphs, feature aggregation, and message-passing techniques.
The results suggest that GNNs can be an effective method for aiding in the decision-making process of MILPs.

Finally, \cite{anderson_generative_2021} proposed a weakly-supervised approach for warm-starting gas network problems.
The authors present a model that generates feasible initial solutions which are used to accelerate the MILP solver's convergence.
The results show a 60.5\% decrease in the runtime.

In summary, the literature offers promising approaches for accelerating MILP solvers using ML techniques, particularly through early fixing.
\new{However, to the best of our knowledge, no one has applied such techniques to oil production optimization.}
In this context, this paper \new{is the first to explore} the use of supervised and weakly-supervised learning approaches to the oil production maximization problem, leveraging surrogate models for the liquid production and offering insights into this growing area that hybridizes optimization and deep learning.

\section{Problem Statement}

In this section, we present \new{the problem formulation, which is based on} \cite{muller_short-term_2022}, with only gas–lifted oil wells connected to manifolds.
Nevertheless, it is easy to see that our methodological approach can be applied to variations of this problem, e.g., platforms with satellite wells, wells with electrical submersible pump systems, and subsea manifolds.

\subsection{Well model}

A single production platform can extract from multiple oil wells.
Each well $n \in \mathcal{N}$ has its liquid production $q_{\rm l}^n$ induced by the wellhead pressure ${\rm whp}^n$ and a lift-gas injection rate $q_{\rm gl}^n$.
The relationship between $q_{\rm l}^n$, ${\rm whp}^n$, and $q_{\rm gl}^n$ is modeled based on the natural characteristics of the well and the gas-oil ratio (GOR, referred to as ${\rm gor}^n$) and basic-sediment-to-water ratio (BSW, referred to as ${\rm bsw}^n$) of its liquid production.
Both ${\rm bsw}^n$ and ${\rm gor}^n$ are measured through separation tests and are considered static during the optimization.
However, as they change with time, their updates drive new executions of the optimization process to keep the results reliable.

\subsection{Piecewise linearization}

We use the {\scshape Marlim} simulator \citep{seman_derivative-free_2020}, a proprietary software from Petrobras, to model the liquid output of each well,
\begin{equation}\label{eq:marlim}
    q_{\rm l}^n = \textrm{\scshape Marlim}(q_{\rm gl}^n, {\rm whp}^n, {\rm bsw}^n, {\rm gor}^n)
.\end{equation}
An example of the liquid flow function of a real well can be seen in Figure \ref{fig:q-liq-surface} with fixed values for ${\rm bsw}^n$ and ${\rm gor}^n$.
As both $q_{\rm gl}^n$ and ${\rm whp}^n$ can be controlled, but have a nonlinear relationship with the outcome, we apply piecewise linearization to $q_{\rm l}^n$ as a function of both.
More precisely, let $\mathcal{K}_{\rm gl} = \{1,\ldots,k_{\rm gl}\}$ and $\mathcal{J}_{\rm whp} = \{1,\ldots,j_{\rm whp}\}$ be sets of indices for lift-gas injection and wellhead pressure.
Let also $\mathcal{Q}_{\rm gl}^{n} =\{q_{\rm gl}^{n,k} \in \mathbb{R}_+ :  k\in \mathcal{K}_{\rm gl}\}$ and $\mathcal{W}^{n}_{\rm whp} =\{{\rm whp}^{n,j} \in \mathbb{R}_+ :  j\in \mathcal{J}_{\rm whp}\}$ be the respective breakpoint values for well $n$, and 
\begin{multline}\label{eq:Q-liq}
  \mathcal{Q}_{\rm liq}^{n} =\left \{ q_{\rm l}^{n,k,j} :   q_{\rm l}^{n,k,j} = \hat{q}_{\rm l}^{n}(q_{\rm gl}^{n,k},{\rm whp}^{n,j},{\rm bsw}^n,{\rm gor}^n), \right .  \\ 
    \left .  {k\in \mathcal{K}_{\rm gl}, j\in \mathcal{J}_{\rm whp}} \right\}
\end{multline}
be the liquid flow rates for the well at the breakpoints.
$\mathcal{Q}_{\rm liq}$ is obtained by using {\scshape Marlim} as in Eq. \eqref{eq:marlim} with the adjusted parameters ${\rm gor}^n$ and ${\rm bsw}^n$.

The piecewise approximation is then given by
\begin{subequations}\label{eq:pwl-approximation}
\begin{align}
    q_{\rm gl}^n ~ & = ~\sum_{k\in \mathcal{K}_{\rm gl}} \sum_{j\in \mathcal{J}_{\rm whp}} \theta^{n}_{k,j} q_{\rm gl}^{n, k} \\
    {\rm whp}^n ~&  = ~\sum_{k\in \mathcal{K}_{\rm gl}} \sum_{j\in \mathcal{J}_{\rm whp}} \theta^{n}_{k,j} {\rm whp}^{n, j} \\
    q_{\rm l}^{n} ~& = ~ \sum_{k\in \mathcal{K}_{\rm gl}} \sum_{j\in \mathcal{J}_{\rm whp}} \theta^{n}_{k,j} q_{\rm l}^{n, k, j}  \\
   1 ~& = ~ \sum_{k\in \mathcal{K}_{\rm gl}} \sum_{j\in \mathcal{J}_{\rm whp}} \theta^{n}_{k,j} \\
   \theta^{n}_{k,j} ~&\geq 0, ~ k\in \mathcal{K}_{\rm gl}, \, j\in \mathcal{J}_{\rm whp}  
.\end{align}
\end{subequations}

To ensure piecewise linearization, SOS2 constraints are necessary for the values of $\theta$ that correspond to $q_{\rm gl}$ and ${\rm whp}$, as follows
\begin{subequations}\label{eq:pwl-consistency}
\begin{align}
    \eta_{k}^{n} ~&=~\sum_{j\in \mathcal{J}_{\rm whp}} \theta^{n}_{k,j}, \, k\in \mathcal{K}_{\rm gl} \\
    \eta_{j}^{n} ~&=~ \sum_{k\in \mathcal{K}_{\rm gl}} \theta^{n}_{k,j}, \, j\in \mathcal{J}_{\rm whp} \\
    &\left \{ \eta_{k}^{n}  \right \}_{ k\in \mathcal{K}_{\rm gl}}  ~~\text{is SOS2} \label{eq:pwl-consistency:SOS2-gl}\\
    &\left \{ \eta_{j}^{n}  \right \}_{ j\in \mathcal{J}_{\rm whp}}  \text{is SOS2} \label{eq:pwl-consistency:SOS2-whp} 
.\end{align}
\end{subequations}
SOS2 constraints imply that no more than two consecutive elements of the ordered set of variables are nonzero \citep{beale_special_1969}.

\subsection{Problem formulation}

All oil wells are connected to a hub that directs their liquid productions $q_{\rm l}^n$ to separators.
The separators then split the liquid flow into the oil, gas, and water phase flows.
This separation depends on the \new{BSW and GOR of each well $n\in \mathcal{N}$,}
\begin{subequations}\label{eq:separator}
\begin{align}
    q_{\rm oil}^n &= q_{\rm l}^n \cdot (1 - {\rm bsw}^n) \\
    q_{\rm water}^n &= q_{\rm l}^n \cdot {\rm bsw}^n \\
    q_{\rm gas}^n &= q_{\rm l}^n \cdot (1 - {\rm bsw}^n) \cdot {\rm gor}^n
.\end{align}
\end{subequations}
The total gas flow is limited by the maximum lift-gas flow available $\bar{q}_{\rm gl}$ though
\begin{equation}\label{eq:q-gl-max}
    \sum_{n\in \mathcal{N}} q_{\rm gl}^n \le \bar{q}_{\rm gl}
.\end{equation}

Finally, the objective is to maximize the total volume of oil extracted
\begin{equation}\label{eq:objective}
    \sum_{n \in \mathcal{N}} q_{\rm oil}^n
.\end{equation}

We can express the problem as
\begin{equation}
\begin{split}
    \max_{\boldsymbol{q}_{\rm gl}, \boldsymbol{{\rm whp}}} 
 ~& \eqref{eq:objective} \\
    \textrm{s.t.} ~~&  \eqref{eq:pwl-approximation} \textrm{--} \eqref{eq:q-gl-max}
,\end{split}
\end{equation}
\new{where $\boldsymbol{q}_{\rm gl}, \boldsymbol{{\rm whp}} \in \mathbb{R}^{|\mathcal{N}|}$ are the flow of lift-gas injected and the well-head pressure of each well.}

Note that the problem, in this case using piecewise linearization of the liquid flow, is an MILP parameterized by ${\rm bsw}^n$, ${\rm gor}^n$, $\bar{q}_{\rm gl}$, and $Q_{\rm liq}^n$.
Therefore, let $\pi \in \Pi$ be the vector of problem parameters.
We can write the problem as
\begin{equation}\label{eq:can-form}
\begin{split}
    P(\pi) = \max_{x, z} ~& c^T x  \\
    \text{s.t.}  ~& A(\pi)x + C(\pi)z \le b(\pi)  \\
    & x \in X \subset \mathbb{R}^{n_x}, z \in Z \subset \{0,1\}^{n_z}
\end{split}
\end{equation}
where $x$ is the vector of continuous variables (e.g., $q_{\rm l}^n$, $q_{\rm gl}^n$, $\theta^n_{k,j}$) and $z$ is the vector of binary variables necessary for the SOS2 constraints.
More precisely, $z = (\boldsymbol{z}_{\rm gl}, \boldsymbol{z}_{{\rm whp}})$ is such that, for all $n\in \mathcal{N}$,
\begin{subequations} \label{eq:SOS2}
\begin{align}
    \eta^n_k &\le z_{\rm gl}^{n,k}, k=1\\
    \eta^n_j &\le z_{{\rm whp}}^{n,j}, j=1 \\
    \eta^n_k &\le \max(z_{\rm gl}^{n,k}, z_{\rm gl}^{n,k-1}), k\in\mathcal{K}_{\rm gl}\setminus \{1, k_{\rm gl}\} \\
       \label{eq:SOS2:max-gl} 
    \eta^n_j &\le \max(z_{{\rm whp}}^{n,j}, z_{{\rm whp}}^{n,j-1}), j\in\mathcal{J}_{{\rm whp}}\setminus \{1, j_{{\rm whp}}\} \\
        \label{eq:SOS2:max-whp} 
    \eta^n_k &\le z_{\rm gl}^{n,k-1}, k=k_{\rm gl} \\
    \eta^n_j &\le z_{{\rm whp}}^{n,j-1}, j=j_{{\rm whp}} \\
    &\sum_{k\in \mathcal{K}_{\rm gl}\setminus\{k_{\rm gl}\}}  z_{\rm gl}^{n,k} = 1, \sum_{j\in \mathcal{J}_{{\rm whp}}\setminus\{j_{{\rm whp}}\}} z_{{\rm whp}}^{n,j} = 1 \\
    &z_{\rm gl}^{n} \in \{0,1\}^{k_{\rm gl}-1} \\
    &z_{{\rm whp}}^{n} \in \{0,1\}^{j_{{\rm whp}}-1}
\end{align}
\end{subequations}
in which $z_{\rm gl}^{n,k}=1$ if the $k$-th interval is selected, meaning that $q_{\rm gl}^n\in[q_{\rm gl}^{n,k},q_{\rm gl}^{n,k+1}]$ and only $\eta^k_n$ and $\eta^{k+1}_n$ can be nonzero, otherwise $z_{\rm gl}^{n,k}$ assumes value $0$. The semantics of the binary variables $z_{\rm whp}^{n,j}$ are analogous. 
Notice that, because $z_{\rm gl}^{n,k}$ is binary, the operator $\max(z_{\rm gl}^{n,k}, z_{\rm gl}^{n,k-1})$ in \eqref{eq:SOS2:max-gl} can be equivalently represented by the linear form $(z_{\rm gl}^{n,k} + z_{\rm gl}^{n,k-1})$.
The formulation \eqref{eq:SOS2} is equivalent to the SOS2 constraints in \eqref{eq:pwl-consistency:SOS2-gl} and \eqref{eq:pwl-consistency:SOS2-whp}.

\section{Methodology}

\subsection{Early fixing}

Suppose we can determine which linearization region of $q_{\rm l}^n$ is to be selected (i.e., which variables $\eta_k^n$ and $\eta_j^n$ can have nonzero values). In that case, the SOS2 constraints can be removed; thus, the problem becomes completely linear.
This is equivalent to fixing the $z$ variables in the standard formulation \eqref{eq:can-form}.
Therefore, let us write $P(\pi,z)$ as the problem \eqref{eq:can-form} but with fixed $z$ values, i.e., with integer variables $z$ treated as parameters.

An early fixing heuristic provides an assignment $\hat{z}$ to the integer variables.
Ideally, the assignment will be such that $P(\pi, \hat{z}) = P(\pi)$.
Since the early-fixed problem is an LP, it can be solved much faster than the original MILP problem.
Therefore, the total cost of solving the early-fixed problem is the cost of solving the LP and the cost of running the heuristic.
In practice, however, a trade-off between the cost of running the heuristic and the gap between $P(\pi,\hat{z})$ and $P(\pi)$ is expected.

Our proposed approach is to develop an early fixing heuristic based on a deep-learning model.
We want a deep learning model $N:\Pi \to [0,1]^{n_z}$ with which we can compute $\hat{z} = \lfloor N(\pi) \rceil$.
Two distinct approaches to training such a model are proposed.

\subsection{Supervised learning approach}

We can train a model for early fixing by feeding it with instances of the MILP problem and optimizing the model’s parameters such that its output approximates the optimal binary assignment.
Let us define a dataset \[
    D_{\rm sup} = \left\{(\pi,z^\star) \in \Pi \times Z: P(\pi, z^\star) = P(\pi)\right\}
,\]
which associates instances' parameter vectors $\pi$ with the optimal binary assignment $z^\star$.
Note that this dataset requires us to solve to optimality all MILP instances available.

Let us define a deep learning model
\begin{equation}
\begin{split}
N_{\rm sup} : \Pi \times \Theta_{\rm sup} &\to [0,1]^{n_z} \\
\pi; \theta_{\rm sup} &\mapsto N_{\rm sup}(\pi ; \theta_{\rm sup})
\end{split}
\end{equation}
for which $\theta_{\rm sup}$ is the vector of model parameters that can be trained.
Then, it is possible to optimize the model's parameters such that for each vector of parameters in $D_{\rm sup}$, the model's output approximates $z^\star$.
Namely,
\begin{equation}\label{eq:min-sup}
\min_{\theta_{\rm sup}} \sum_{(\pi, z^\star) \in D_{\rm sup}} \mathcal{L}_{\rm sup}(N_{\rm sup}(\pi ; \theta_{\rm sup}), z^\star)
\end{equation}
where $\mathcal{L}_{\rm sup}$ can be, for example, the binary cross entropy between the elements of both vectors.

\subsection{Weakly-supervised learning approach}\label{sec:weakly-train}

We propose an alternative learning approach that does not require solving MILP problems.
First, we recall that the target of the deep learning model is to provide $\hat{z}$ such that $P(\pi, \hat{z})$ is maximized\footnote{Since \eqref{eq:can-form} is a maximization problem.}.
Our proposed approach is to train a \emph{surrogate} model that approximates $P(\cdot,\cdot)$, and differentiate through this surrogate to train the early fixing heuristic using gradient descent methods.

This approach requires only a dataset of assignments for the integer variables paired with the objective of the respective early-fixed problem.
Let us define \[
D_{\rm weak} = \left\{(\pi,\hat{z},p) \in \Pi \times Z \times \mathbb{R}: p = P(\pi, \hat{z})\right\}
,\]
which is built with (random) samples of $\hat{z} \in Z$.
Then, we train a model
\begin{equation}
\begin{split}
S : \Pi \times [0,1]^{n_z} \times \Theta_S & \to \mathbb{R} \\
\pi, \hat{z}; \theta_S &\mapsto S(\pi, \hat{z} ; \theta_S)
\end{split}
\end{equation}
in a supervised manner, such that its output approximates $P(\pi, \hat{z})$, that is,
\begin{equation}\label{eq:min-S}
    \min_{\theta_S} \sum_{(\pi, \hat{z}, p) \in D_{\rm weak}} \mathcal{L}_S(S(\pi, \hat{z} ; \theta_S), p)
,\end{equation}
where $\mathcal{L}_S$ can be, for example, the mean squared error.

Now, let us define
\begin{equation}
\begin{split}
N_{\rm weak} : \Pi \times \Theta_{\rm weak} &\to [0,1]^{n_z} \\
\pi; \theta_{\rm weak} &\mapsto N_{\rm weak}(\pi ; \theta_{\rm weak})
,\end{split}
\end{equation}
which can be trained in an unsupervised manner to maximize the surrogate model’s output given the candidate values for fixing
\begin{equation}\label{eq:min-weakly}
    \max_{\theta_{\rm weak}} \sum_{\pi \in_R \Pi} S(\pi, N_{\rm weak}(\pi ; \theta_{\rm weak}) ; \theta_S)
,\end{equation}
where $\pi \in_R \Pi$ denotes that $\pi$ is chosen randomly from $\Pi$, and the summation is iterated over as many random samples as desired.
Note that only $\theta_{\rm weak}$ is optimized during the training of $N_{\rm weak}$, i.e., $S$ is unchanged.

\section{Experiments and Results\protect\footnote{\NoCaseChange{Code available in \href{https://github.com/brunompacheco/early-fixing-oil-production}{github/brunompacheco/early-fixing-oil-production}.}}}

\subsection{Data}

For our experiments, we consider the problem of optimizing oil production from a single well ($|\mathcal{N}| = 1$).
We use data from a real subsea oil well, provided by Petrobras\footnote{\new{The data is not made available as it is an intellectual property of Petrobras.}}.
We set a target for the early fixing models to generalize to different values of ${\rm bsw}$, ${\rm gor}$, and $\bar{q}_{\rm gl}$.
Therefore, we describe the parameter space of our problem as \[
    \pi = ({\rm bsw}, {\rm gor}, \bar{q}_{\rm gl}) \in \Pi = [0.5, 1.0] \times [0, 300] \times [4000,12500]
\] assuming that the BSW can be no lower than 0.5, the GOR is always smaller than 300, and the maximum lift-gas flow is always larger than $10^5$.
More precisely, we assume that, in practice, ${\rm bsw} \sim U(0.5, 1.0)$, ${\rm gor} \sim N(100, 25)$, and $\bar{q}_{\rm gl} \sim U(4000,12500)$.
Note that we omit the liquid flow function from the parameter vector, once we deal with a single well and, thus, $\mathcal{Q}_{\rm liq}$ can be uniquely determined by the other parameters.

\begin{figure}
    \centering
    \includegraphics[width=0.5\textwidth]{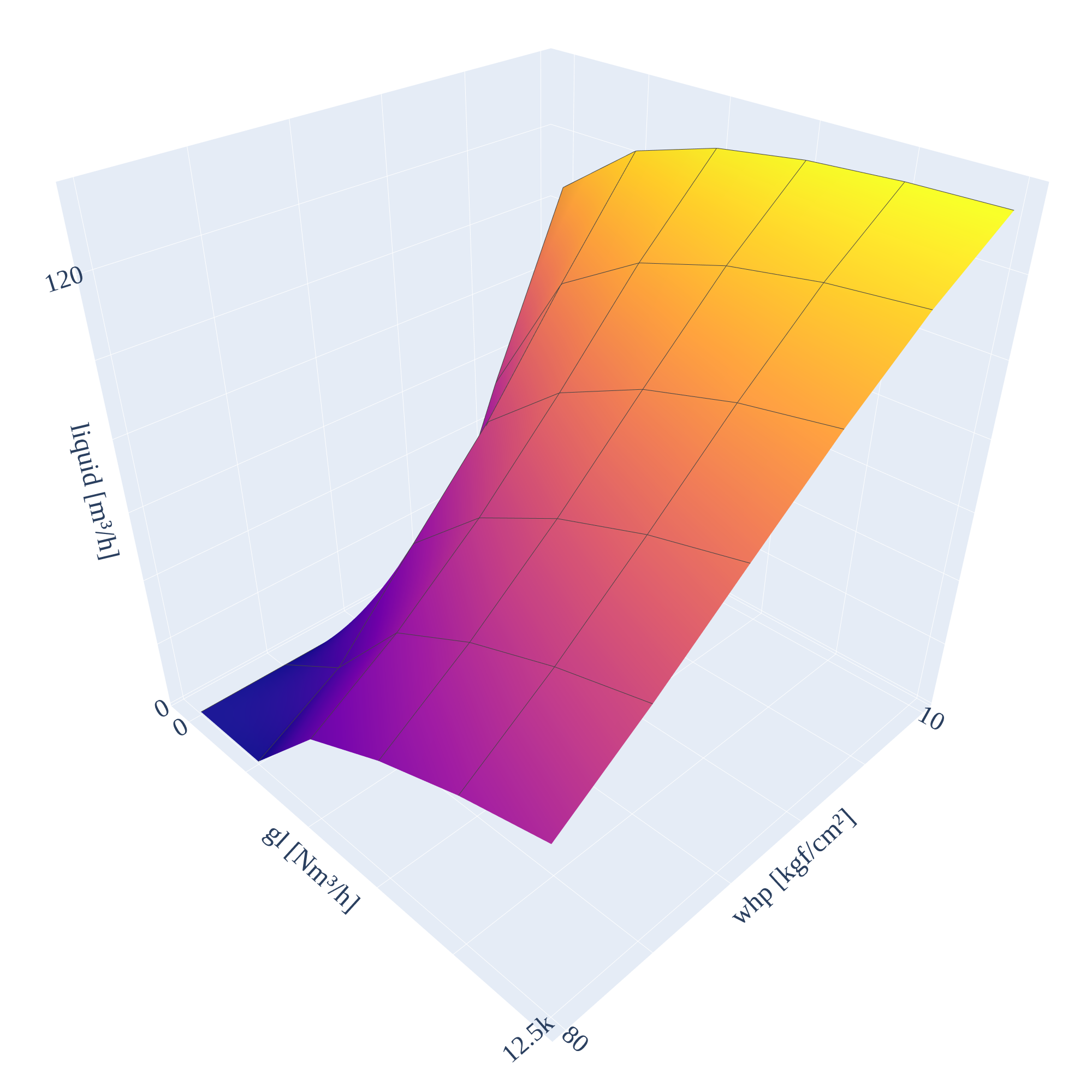}
    \caption{Simulated liquid flow $\mathcal{Q}_{\rm liq}$ (see Eq. \eqref{eq:Q-liq}) based on data of a real well using {\scshape Marlin}. The color scale is directly proportional to the z-axis. We adopted $q_{\rm l}^{n,k,j} = -1$ for  combinations of $q_{\rm gl}^{n,k}$ and ${\rm whp}^{n,j}$ that were considered infeasible by the simulator.}
    \label{fig:q-liq-surface}
\end{figure}

The liquid flow function is always linearized with the same breakpoints for both $q_{\rm gl}$ and ${\rm whp}$.
This makes the domain of the binary variables consistent across instances.
The selected breakpoints are
\begin{subequations} \label{eq:range-qgl-whp}
\begin{align}
    {\rm whp}^{j} &= 14k - 4,\quad &k =1,\dots,6 \\
    q_{\rm gl}^{k} &= 2500j - 2500,\quad &j =1,\dots,6
.\end{align}
\end{subequations}
An example of $\mathcal{Q}_{\rm liq}$ with these breakpoints can be seen in Figure \ref{fig:q-liq-surface}.
With the fixed breakpoints, we can describe the domain of the binary variables as
\begin{equation}
\begin{split}
Z = \left\{(z_{\rm gl}, z_{{\rm whp}}) \in \{0,1\}^5 \times \{0,1\}^5 : \sum_{i=1}^5 z_{\rm gl} = 1,\right. \\
\left. \sum_{i=1}^5 z_{{\rm whp}} = 1\right\}
,\end{split}
\end{equation}
in which $z_{\rm gl}$ indicates the pair of $\eta_k$ variables that can take nonzero values, while $z_{{\rm whp}}$ indicates the pair of $\eta_j$ variables that can take nonzero values.

\new{For the supervised learning task, we build $D_{\rm sup}$ with 500 instances of the MILP problem from different combinations of ${\rm bsw}$, ${\rm gor}$, and $\bar{q}_{\rm gl}$.}
Gurobi was used to solve the MILPs, upon which $z^\star = (z^\star_{\rm gl}, z^\star_{{\rm whp}})$, the optimal value for the binary variables, was extracted.
We use the same $\pi$ vectors as in $D_{\rm sup}$ for the weakly-supervised learning task. For each $\pi$ we draw 12 random candidates $\hat{z} \in Z$, that is, we ensure that each $\hat{z}$ respects the constraints in \eqref{eq:SOS2}.
Then, $D_{\rm weak}$ is built by solving the LPs using Gurobi to compute $p=P(\pi, \hat{z})$.
Whenever $\pi$ and $\hat{z}$ resulted in an infeasible problem, we added them to $D_{\rm weak}$ with $p = -1$.
In total, $D_{\rm sup}$ contains 500 \new{data points} of the form $(({\rm bsw}, {\rm gor}, \bar{q}_{\rm gl}), z^\star)$, while $D_{\rm weak}$ contains 6000 data points of the form $(({\rm bsw}, {\rm gor}, \bar{q}_{\rm gl}), \hat{z})$.

\subsection{Supervised learning experiments}\label{sec:sup-learning-experiments}

For supervised learning, using $D_{\rm sup}$, we choose
\begin{equation}
\begin{split}
N_{\rm sup} : \Pi \times \Theta_{\rm sup} &\to \Delta^5 \times \Delta^5 \\
{\rm bsw}, {\rm gor}, \bar{q}_{\rm gl}; \theta_{\rm sup} &\mapsto (\hat{z}_{\rm gl}, \hat{z}_{{\rm whp}}) = N_{\rm sup}({\rm bsw}, {\rm gor}, \bar{q}_{\rm gl} ; \theta_{\rm sup})
,\end{split}
\end{equation}
where $\Delta^5$ is the $5$-dimensional simplex set, as a neural network with 2 hidden layers of $25$ neurons each and a $10$-dimensional output.
The inputs are normalized before the first layer.
We use ReLU activation for all layers except the last one.
The last layer’s output is divided into two vectors of dimension $5$, passing through the softmax function, thus mapping both into 5-dimensional simplexes.
The softmax in the output of the network ensures that, after rounding, it respects the binary constraints, that is, $\lfloor \hat{z}_{\rm gl} \rceil$ and $\lfloor \hat{z}_{{\rm whp}} \rceil$ respect \eqref{eq:SOS2}.

We randomly split $D_{\rm sup}$ into training and test sets with an 80-20 ratio.
\new{%
The training data was used to select the optimal network architecture and hyperparameters (learning rate, batch size, and number of epochs).
After this adjustment, the entirety of the training set was used to train} $5$ $N_{\rm sup}$ models with random initialization.
Adam is used to optimizing the parameters of the models on the training set such that $\mathcal{L}_{\rm sup}$ is minimized (see equation \eqref{eq:min-sup}).
We use
\begin{multline*}
    \mathcal{L}_{\rm sup}(\hat{z}_{\rm gl}, \hat{z}_{{\rm whp}}, z_{\rm gl}^\star, z_{{\rm whp}}^\star) = \sum_{i=1}^{k_{\rm gl}-1}{\rm BCE}(\hat{z}_{{\rm gl},i}, z_{{\rm gl},i}^\star) \\
    + \sum_{j=1}^{k_{\rm whp}-1}{\rm BCE}(\hat{z}_{{\rm whp},j}, z_{{\rm whp},j}^\star)
,\end{multline*}
where BCE is the binary cross-entropy function.
We use batches of $64$ elements and an initial learning rate of 0.001.
Each model is trained for $100$ epochs.
The performance on the test set is summarized in Table \ref{tab:test-results}.

\begin{table}[hb]
\begin{center}
\caption{Early fixing performance on the test set. We measure the models' accuracy in predicting the binary variables' optimal values. \emph{Infeasible} is the ratio of instances that became infeasible after early fixing. \emph{Objective gap} is the mean relative decrease of objective value by performing early fixing, for all instances that remained feasible after early fixing. The values reported are the average of 5 runs (models with random initialization), except for \emph{Baseline}. }\label{tab:test-results}
\begin{tabular}{lccc}
\toprule
Model & Accuracy & Infeasible & Objective gap \\
\midrule
Supervised                & 99.78\%  & 0.11\%      & 0.01\%              \\
Weakly-supervised           & 32.31\%  & 4.19\%       & 3.64\%              \\
Baseline                  & 0\%      & 0\%          & 21.42\%            \\
\bottomrule
\end{tabular}
\end{center}
\end{table}

\subsection{Weakly-supervised learning experiments}

We build the surrogate model
\begin{equation}
\begin{split}
S : \Pi \times [0,1]^{10} \times \Theta_S & \to \mathbb{R} \\
{\rm bsw}, {\rm gor}, \bar{q}_{\rm gl}, z_{\rm gl}, z_{{\rm whp}}; \theta_{S} &\mapsto \hat{p} = S({\rm bsw}, {\rm gor}, \bar{q}_{\rm gl}, z ; \theta_{S})
\end{split}
\end{equation}
as a neural network with $3$ hidden layers of $10$ neurons each.
ReLU is used as an activation function at each layer except the last one, which has no activation function.
Inputs are normalized and a factor of $2000$ scales the output.
Dropout is applied during training at each hidden layer with a probability of $20$\% for each neuron.

\begin{figure}
    \centering
    \includegraphics[width=0.5\textwidth]{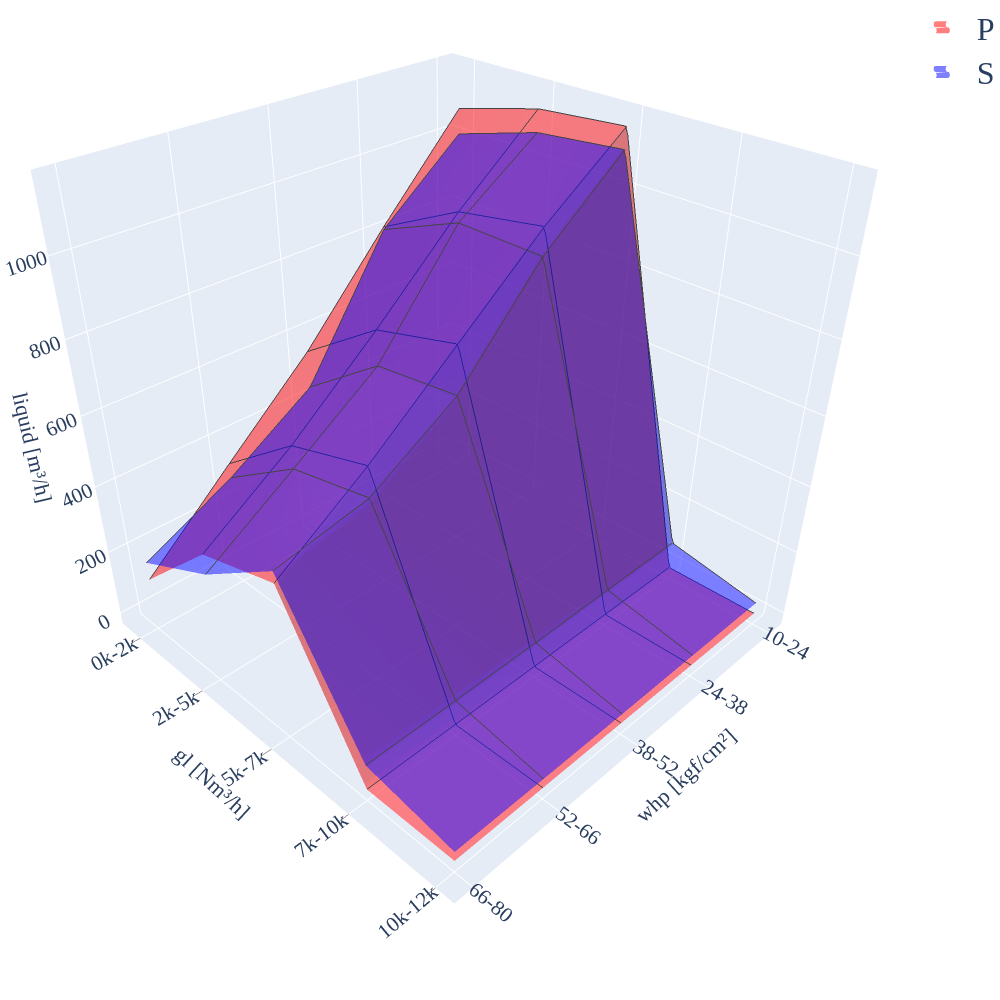}
    \caption{Comparison of the surrogate model $S(\pi,z)$ (in blue) and the target function $P(\pi,z)$ (in red) for all early fixing possibilities ($\forall z \in Z$) for a sample instance of the problem. Values of -1 in \emph{P} indicate that the problem is infeasible for that given combination of the variables.}
    \label{fig:P-S-surfaces}
\end{figure}

The dataset $D_{\rm weak}$ is split randomly into a training and test set following an 80-20 ratio.
\new{The optimal architecture and hyperparameters (learning rate, batch size, number of epochs) were determined in the same way as in Sec. \ref{sec:sup-learning-experiments}.}
We use the squared error as the loss function \[
    \mathcal{L}_S(\hat{p}, p) = |\hat{p} - p|^2
.\]
We use Adam to minimize the loss function on the training set, with an initial learning rate of $0.001$ and mini-batches of $1024$ samples.
$5$ models with random initialization are trained for $20$ epochs each.
On the test set, the surrogate models correctly predicted the feasibility (its output was lower than $0$) an average of $78.23$\% of the time.
The average MAE on the instances that were not infeasible was $60.05$.
Figure \ref{fig:P-S-surfaces} shows an example of the surrogate model's performance compared to the real objective values.

The early fixing model $N_{\rm weak}$ has the same architecture as $N_{\rm sup}$.
$5$ models with random initialization are trained as described in Section \ref{sec:weakly-train}, but using the parameters $\pi$ from the training set of $D_{\rm sup}$ described above.
Each $N_{\rm weak}$ model is trained with a different surrogate model $S$.
The models are trained using Adam with an initial learning rate of $0.01$ and a batch of $64$ for $100$ epochs each.
The performance on the $D_{\rm sup}$ test set is reported in Table \ref{tab:test-results}.

\subsection{Baseline Model}

As a reference for the deep learning models’ performance, we compute the results of always fixing the same values for the binary variables.
In our problem, the safest option in this approach is always to pick the region with the smallest values possible for $q_{\rm gl}$ and ${\rm whp}$.
This always results in a feasible problem.
The performance of this baseline approach on the test set of $D_{\rm sup}$ can be seen in Table \ref{tab:test-results}.

\subsection{Early Fixing Impacts}

To evaluate the impacts of early fixing in the optimization, we measure the runtime of solving the original MILP, the runtime of the early-fixed problem (which is an LP), and the runtime of the early fixing models.
We perform these experiments on all instances of $D_{\rm sup}$, i.e., with the problems defined by the parameters $\pi$ in the $D_{\rm sup}$ dataset used in the experiments above.

We found that the original problem can be solved, on average, in 0.90\,ms, while the early-fixed problem is solved in an average of 0.18\,ms.
Considering the 0.08\,ms the early-fixing models took, on average, during the experiments, the early-fixing approach takes, on average, 0.26\,ms, representing a 71.11\% runtime reduction.

\section{Conclusions}

Our experiments show that deep-learning-based early fixing models successfully speed up the optimization of the offshore gas-lifted oil production problem, with a 71.11\% runtime reduction.
Training in a supervised learning setting, although with a significantly higher cost for collecting the training data, is undoubtedly a superior approach concerning the weakly-supervised setting.
Nevertheless, the experiments with the weakly-supervised approach indicate that it is possible to develop an early fixing heuristic when optimal solutions to the MILPs are unavailable or too hard to obtain.
Still, the weakly-supervised approach needs further refinement to achieve competitive results.

Further research is still necessary on the suitability of the deep-learning-based early fixing for harder problems, e.g., multiple wells connected to manifolds, with more operational constraints and more integer variables.
Moreover, the early fixing approaches presented in this paper theoretically apply to any MILP.
However, the performance of the models and the practical viability of them is still an open theme for research.


\bibliography{references}             


\end{document}